\documentclass[eandd]{article}

\usepackage[eandd, final]{neurips_2026}
\usepackage{amsmath}
\usepackage{amssymb}
\usepackage{booktabs}
\usepackage{multirow}
\usepackage{graphicx}
\usepackage{svg}
\usepackage{subcaption}
\usepackage{wrapfig}
\usepackage[table]{xcolor}
\usepackage{array}
\usepackage{placeins}
\usepackage{longtable}

\usepackage[utf8]{inputenc} 
\usepackage[T1]{fontenc}    
\usepackage{hyperref}       
\usepackage{url}            
\usepackage{booktabs}       
\usepackage{amsfonts}       
\usepackage{nicefrac}       
\usepackage{microtype}      
\usepackage{xcolor}         

\title{HAVEN: Hierarchically Aligned Multimodal Benchmark for Unified Video Understanding}

\author{%
  Mengqi Shi \hfill Haopeng Zhang\thanks{Corresponding author} \\
  Department of Information and Computer Sciences \\
  University of Hawaii at Manoa \\
  \texttt{\{shim7, haopengz\}@hawaii.edu}
}

\begin{document}

\maketitle

\begin{abstract}
While Multimodal Large Language Models (MLLMs) exhibit strong performance on standard video tasks, their ability to faithfully summarize and reason over complex narratives remains poorly evaluated. Existing summarization benchmarks fragment supervision across isolated granularities, such as keyframes, key shots, or disjointed text summaries, failing to capture the inherently hierarchical structure of cross-modal alignment. To address this critical gap, we introduce HAVEN, a hierarchically aligned multimodal benchmark for unified video understanding. HAVEN pioneers a fully granular (frame, shot, and video levels) and fully multimodal (video and text) dataset architecture, complete with explicit, continuous alignment between modalities. Built upon this unified annotation paradigm, we propose a comprehensive evaluation suite spanning summarization, temporal reasoning, multimodal grounding, and saliency ranking. Extensive benchmarking of state-of-the-art MLLMs exposes a persistent gap between surface-level textual fluency and grounded multimodal understanding. Ultimately, HAVEN advances the evaluation of multimodal systems beyond traditional QA formats, offering a rigorous, standardized testbed to drive future research in interpretable, hierarchical video understanding. We publicly release the dataset, benchmark suite, and evaluation protocols at \url{https://github.com/alohalab-ai/align_vsum}.
\end{abstract}

\section{Introduction}
Multimodal large language models (MLLMs) have rapidly advanced video understanding, achieving strong performance on tasks such as summarization~\citep{VideoXum2024lin, MSMO2018zhu, Shot2Story2025han}, captioning~\citep{xu2016msr, wang2019vatex, caba2015activitynet}, and question answering~\citep{MVBench2024li, VideoMME2025fu, TempCompass2024liu}. However, these gains raise a fundamental question: \emph{do MLLMs genuinely understand video content, or do they produce plausible outputs by exploiting language priors?} Resolving this ambiguity is challenging due to the lack of evaluation frameworks capable of rigorously assessing fine-grained, temporal alignment between visual evidence and generated language.

Current video benchmarks suffer from two systemic limitations~\citep{MVBench2024li, VideoMME2025fu, chandrasegaran2024hourvideo, TempCompass2024liu, liu2024mmbench}. First, they provide {fragmented supervision}. Most datasets operate at a single temporal granularity (e.g., keyframes, clips, or video-level summaries), ignoring the inherently hierarchical structure of real-world videos, where frames build shots, and shots construct coherent narratives. Second, they lack {precise cross-modal grounding}. Text annotations are typically weakly aligned with visual content, enabling models to succeed without accurately localizing or reasoning over relevant evidence~\cite{yuan2025understanding}. Together, these limitations create an {evaluation blind spot}: models can achieve high scores through fluent generation while exhibiting poor multimodal alignment. As a result, current evaluation protocols systematically overestimate the capabilities of modern MLLMs, particularly on tasks that do not require temporal compositionality or saliency-aware reasoning~\citep{TempCompass2024liu, MVBench2024li, VideoMME2025fu, li2023evaluating, sun2024aligning}.

To address these limitations, we introduce HAVEN (Hierarchically Aligned multimodal benchmark for unified Video undErstandiNg), a comprehensive dataset and evaluation framework designed to systematically measure \emph{hierarchical cross-modal alignment}. HAVEN is built upon two core principles: (1) \textbf{Hierarchical completeness}, providing unified annotations at the frame, shot, and video levels to enable evaluation across multiple temporal scales; and (2) \textbf{Cross-modal continuity}: dense, explicit alignment between visual content and textual descriptions, ensuring that language must be grounded in temporally localized evidence.

Pioneering a fully granular and fully multimodal dataset architecture, HAVEN supports a suite of tasks spanning summarization, temporal reasoning, multimodal grounding, and saliency ranking, each targeting a distinct aspect of video understanding while sharing a common annotation backbone. Crucially, this enables controlled and fine-grained analysis of model behavior across levels of abstraction and modalities.

\begin{wrapfigure}{r}{0.48\textwidth}
  \centering
  \vspace{-10pt}
  \includegraphics[width=\linewidth]{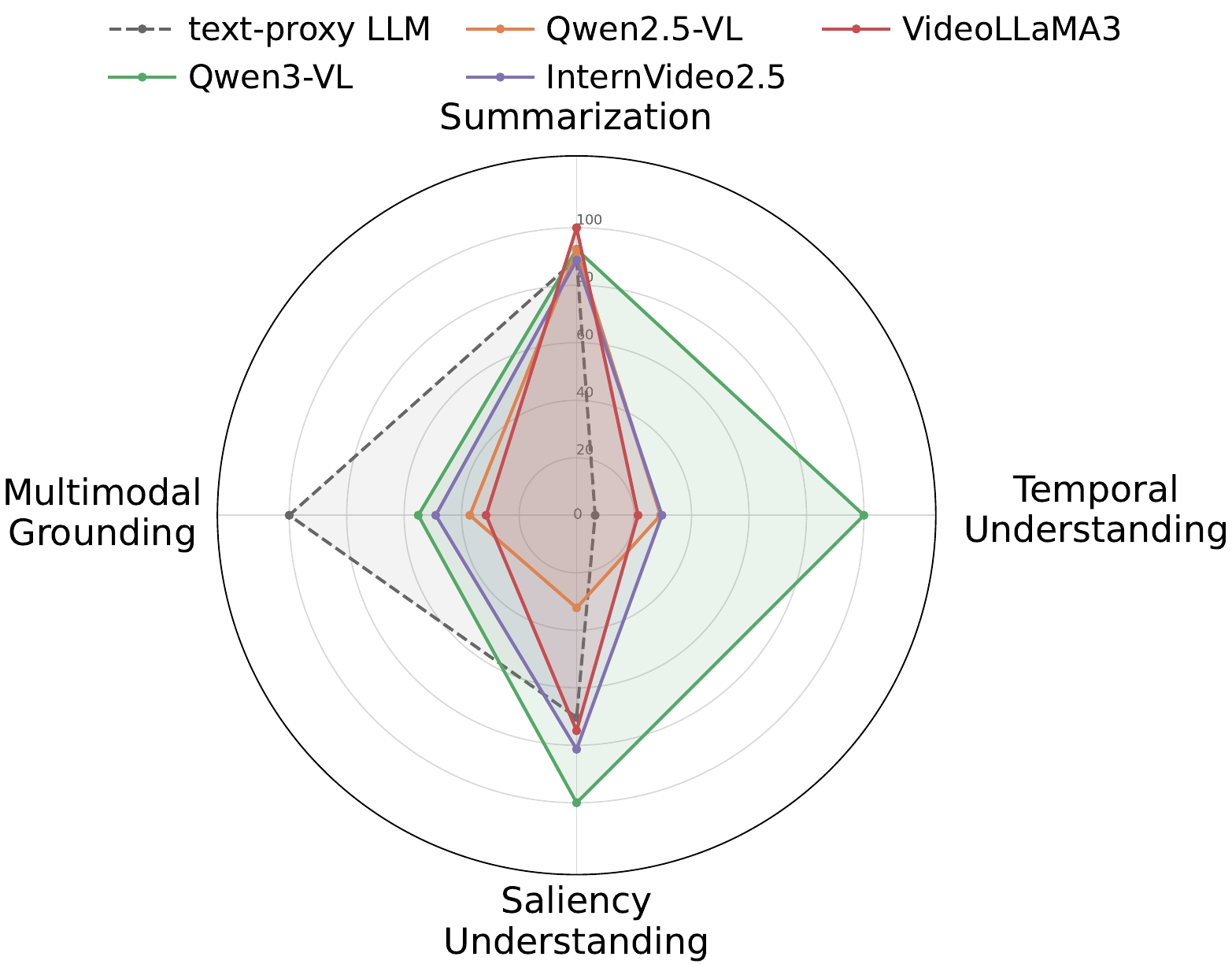}
  \caption{Comparison between different MLLMs on HAVEN across different capabilities. We include text-proxy LLMs as baselines.}
  \label{fig:overall_radar}
\end{wrapfigure}

We benchmark a range of state-of-the-art MLLMs on HAVEN (Fig.~\ref{fig:overall_radar}) and reveal a critical "illusion of capability." While models consistently produce fluent textual summaries, they frequently fail to ground their outputs in the correct visual evidence. Moreover, we observe that joint multimodal generation does not reliably improve alignment, highlighting the fragility of current approaches. These results expose a significant gap between apparent and actual understanding. These findings highlight the significant risk of overestimating MLLM capabilities when evaluations are limited to coarse or weakly grounded tasks.

Our main contributions are three-fold: (1) We introduce HAVEN, a novel dataset that pioneers a fully granular (frame-, shot-, and video-level) and fully multimodal framework with explicit, continuous cross-modal alignment.
(2) We establish a rigorous testing framework featuring distinct tasks across four core dimensions (summarization, multimodal grounding, temporal reasoning, and saliency ranking), enabling a multi-faceted assessment of model capabilities.
(3) We provide an extensive empirical analysis that identifies key limitations in state-of-the-art MLLMs, exposing a severe discrepancy between surface-level textual fluency and true multimodal grounding.

\section{Related Work}

\paragraph{Video Understanding Benchmarks.}
Recent benchmarks have advanced the evaluation of MLLMs on video understanding, covering tasks such as event understanding, temporal reasoning, question answering, and multi-shot comprehension ~\citep{MVBench2024li,VideoMME2025fu,TempCompass2024liu,Shot2Story2025han, caba2015activitynet, huang2025autonomous}. Recent works have introduced benchmarks designed for long video understanding, such as Hourvideo~\citep{chandrasegaran2024hourvideo} and other Q\&A-based benchmarks\citep{xiao2021nextqa,xiao2024nextgqa}.
These benchmarks extend evaluation to more realistic long-video scenarios, but are typically organized around isolated task formats and focus on general understanding. As a result, they provide limited support for jointly evaluating model capabilities of diverse dimensions within a unified framework. Compared to existing benchmarks, we enhance the dataset with richer multimodal signals and fine-grained cross-modal alignment, which enables a more diverse set of tasks beyond traditional Q\&A or coarse-grained formulations. This provides a new perspective for evaluating video understanding.

\paragraph{Video Summarization Benchmarks.}
Video summarization research has been supported by datasets such as SumMe~\citep{Creating2014gygli}, TVSum~\citep{TVSum2015yalesong}, and subsequent resources ~\citep{sul2023mr,VideoXum2024lin}, enabling studies on key segment selection, importance estimation, and summary generation ~\citep{Diverse2014gong,zhang2016video,Video2019rochan}. Other large-scale video-text datasets ~\citep{wang2023internvid,chen2024panda,huang2020movienet,zhang2025bridging} support multiple tasks on video understanding evaluations. More recent work has explored multimodal understanding and evaluation, including cross-modal integration, factual consistency, and evaluation reliability ~\citep{MSMO2018zhu,MDSEval2025liu,Rethinking2019otani,wan2022evaluating, li2026mmvir}. In addition, prior studies highlight the importance of alignment and structured representations for understanding long or segmented content~\citep{Power2024ernst,Unifying2022cho}. However, existing datasets and evaluation protocols are typically designed for single granularity, and lack hierarchical annotations and explicit cross-modal alignments, making it difficult to evaluate whether models can connect textual outputs with their supporting visual evidence across tasks.
\section{HAVEN Benchmark}

\begin{table*}[t]
\caption{\textbf{Source datasets used to construct HAVEN Benchmark.}. ``Alignment`` denotes aligned annotations between visual and texual data, ``$K_{f}$'' = key frames, ``$D_{f}$'' = frame descriptions, ``$K_{s}$'' = key shots, and ``$D_{s}$'' = shot descriptions. ``Avg Frames'' and ``Avg Shots'' denote the average sampled frames and shots per video.}
\centering
\small
\setlength{\tabcolsep}{4pt}
\renewcommand{\arraystretch}{1.05}
\begin{tabular}{lccccccccc}
\hline
\textbf{Dataset} & \textbf{\#Videos} & \textbf{Modalities} & \textbf{Alignment} & \textbf{$K_{f}$} & \textbf{$D_{f}$} & \textbf{$K_{s}$} & \textbf{$D_{s}$} & \textbf{Avg Frames} & \textbf{Avg Shots} \\
\hline
\rowcolor{blue!6}\textbf{HAVEN} & \textbf{11055} & \textbf{V+T$\rightarrow$V+T} & \textbf{$\checkmark$} & \textbf{$\checkmark$} & \textbf{$\checkmark$} & \textbf{$\checkmark$} & \textbf{$\checkmark$} & \textbf{164.9} & \textbf{16.2} \\
\hline
$\hookrightarrow$ SumMe (\citeyear{Creating2014gygli})       & 25    & V$\rightarrow$V     & $\times$      & $\checkmark$ & $\times$      & $\checkmark$ & $\times$ & 293.4 & 29.8 \\
$\hookrightarrow$ TVSum (\citeyear{TVSum2015yalesong})        & 50    & V$\rightarrow$V     & $\times$      & $\checkmark$ & $\times$      & $\checkmark$ & $\times$ & 470.2 & 47.5 \\
$\hookrightarrow$ OVP (\citeyear{de2011vsumm})          & 50    & V$\rightarrow$V     & $\times$      & $\checkmark$ & $\times$      & $\times$     & $\times$ & 103.0 & 9.9 \\
$\hookrightarrow$ YouTube (\citeyear{de2011vsumm})      & 40    & V$\rightarrow$V     & $\times$      & $\checkmark$ & $\times$      & $\times$     & $\times$ & 409.8 & 40.5 \\
$\hookrightarrow$ VideoXum (\citeyear{VideoXum2024lin})     & 5300  & V$\rightarrow$V     & $\times$      & $\checkmark$ & $\times$      & $\checkmark$ & $\times$ & 124.0 & 12.0 \\
$\hookrightarrow$ MR.HiSum (\citeyear{sul2023mr})     & 5590  & V+T$\rightarrow$V   & $\checkmark$  & $\checkmark$ & $\times$      & $\checkmark$ & $\times$ & 202.5 & 19.8 \\
\hline
\end{tabular}
\label{tab:source_dataset_overview}
\end{table*}

\begin{figure}[b]
  \centering
  \vspace{-10pt}
  \includegraphics[width=
  \linewidth]{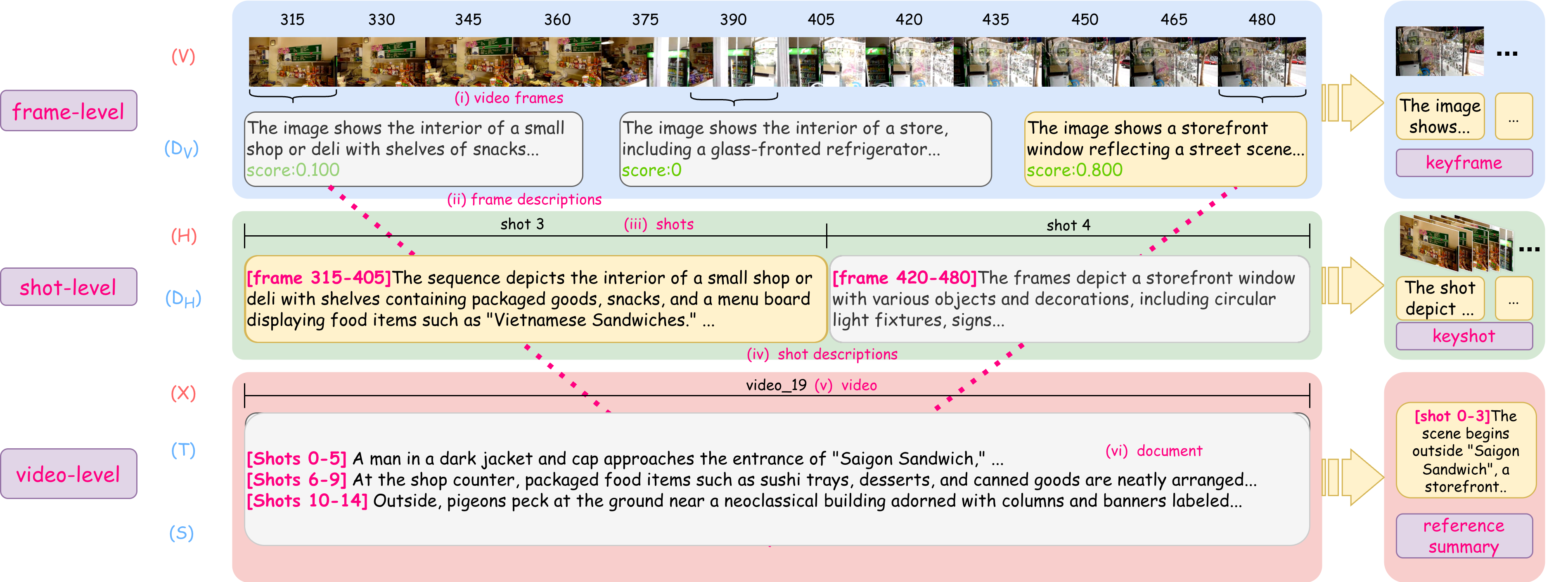}
  \caption{Example of a data instance in HAVEN. We construct a hierarchically structured, multimodal dataset with aligned annotations at the frame, shot, and video (document) levels. Frame- and shot-level textual descriptions are generated from visual inputs and refined for quality, while higher-level document and summary texts are composed by aggregating shot-level descriptions. Human annotations provide ground-truth signals for keyframes and key shots.}
  \label{fig:dataset_construction}
\end{figure}

HAVEN is a novel benchmark that introduces a fully granular (frame-, shot-, and video-level) and fully multimodal framework with explicit, continuous cross-modal alignment. It is built upon six video summarization datasets (see Table~\ref{tab:source_dataset_overview}), with textual components generated by LLMs and further refined for quality. These datasets are selected for their high-quality human-annotated keyframes, which enable the derivation of frame-level importance scores and support hierarchical annotation construction. Each instance corresponds to a single video and includes both visual and textual modalities with aligned hierarchical annotations.

\subsection{Data Instance Representation}

As illustrated in Fig.~\ref{fig:dataset_construction}, each data instance in HAVEN comprises hierarchically structured visual and textual modalities at the frame, shot, and video levels.

At the \textbf{frame level}, a video is represented by a set of sampled frames $V$. Each frame is paired with a textual description $D_V$ that provides fine-grained semantic details, along with a subset of human-annotated keyframes $V_{\text{key}}$.

At the \textbf{shot level}, temporally contiguous frames are grouped into a set of shots $H$, where each shot corresponds to a subset of frames from $V$. Each shot is associated with a shot-level description $D_H$, capturing higher-level semantics and interactions across frames, as well as a subset of human-annotated key shots $H_{\text{key}}$.

At the \textbf{video level}, each video is paired with a document $T = \{t_j\}_{j=1}^{M}$ that provides a global description, and a summary $S = \{s_j\}_{j=1}^{P}$ that captures the key content of the video.

This hierarchical representation establishes explicit multi-level alignments between visual and textual modalities, enabling unified and consistent evaluation across tasks. An example of these hierarchical multimodal annotations is provided in Appendix~\ref{app:dataset_sample}.

\begin{table}[t]
\caption{Human evaluation of annotation quality.}
\centering
\footnotesize
\setlength{\tabcolsep}{5pt}
\renewcommand{\arraystretch}{0.98}
\begin{tabular}{llcccc}
\hline
\textbf{Level} & \textbf{Metric} & \textbf{Score} & $\kappa_w$ & $\rho$ & \textbf{Exact} \\
\hline
Frame & Accuracy & 4.42 $\pm$ 0.39 & -0.15 & -0.11 & 43.3\% \\
 & Informativeness & 4.20 $\pm$ 0.60 & 0.17 & 0.01 & 26.7\% \\
\hline
Shot & Accuracy & 4.43 $\pm$ 0.32 & -0.48 & -0.71 & 13.3\% \\
 & Informativeness & 4.30 $\pm$ 0.32 & -0.58 & -0.66 & 6.7\% \\
\hline
Document & Accuracy & 4.47 $\pm$ 0.35 & -0.20 & -0.21 & 40.0\% \\
 & Completeness & 4.60 $\pm$ 0.34 & -0.30 & -0.30 & 40.0\% \\
 & Readability & 3.97 $\pm$ 0.40 & -0.55 & -0.56 & 13.3\% \\
\hline
Summary & Accuracy & 4.33 $\pm$ 0.49 & 0.26 & 0.29 & 46.7\% \\
 & Relevance & 4.47 $\pm$ 0.48 & -0.12 & -0.11 & 40.0\% \\
 & Conciseness & 4.07 $\pm$ 0.42 & 0.00 & 0.00 & 53.3\% \\
 & Readability & 4.10 $\pm$ 0.54 & -0.07 & -0.14 & 46.7\% \\
\hline
\end{tabular}
\label{tab:human_eval_mms}
\end{table}

\subsection{Construction Process}

The data instances in HAVEN contain aligned visual and textual information across three hierarchical levels:

\textbf{Frame-level.}
We uniformly sample frames from each video at 1–2 FPS, following the conventions of prior datasets~\citep{de2011vsumm, zhang2016video, sul2023mr, VideoXum2024lin}, resulting in an average of \textit{164.9} frames per video. Each video includes \textsc{user-annotated} keyframes from the original datasets, which we convert into frame-level importance scores. The top 15\% of frames~\citep{zhang2016video} are selected as keyframes, serving as ground truth for keyframe selection. Additionally, each sampled frame is paired with a textual description $D_V$, generated by an LLM and refined through human verification.

\textbf{Shot-level.}
Following~\citet{zhang2016video}, we partition each video into temporally coherent shots $H$ using Kernel Temporal Segmentation (KTS), targeting approximately $0.1 \times |V|$ segments. Each shot is associated with a description $D_H$, generated by providing all frames within the shot to an MLLM under a length constraint to ensure concise yet informative summaries. Shot-level ground truth is derived by aggregating frame-level importance scores, after which key shots are selected via a knapsack algorithm with a budget constraint such that the total number of frames in selected shots does not exceed 15\% of $|V|$. This level supports temporal understanding, multimodal grounding, and saliency ranking tasks.

\textbf{Video-level.} 
At the highest level, we construct a document $T = \{t_j\}_{j=1}^{M}$ by providing all shot-level descriptions to an LLM, which groups semantically related and temporally consistent shots into coherent sentences. This process ensures a fluent narrative structure while preserving detailed content. A summary $S = \{s_j\}_{j=1}^{P}$ is constructed similarly, using only key-shot descriptions. In both cases, sentences are aligned with their corresponding shot groups, establishing explicit links between textual units and visual segments. These representations serve both as inputs for certain tasks and as reference text for evaluation. More details on the dataset construction pipeline and annotation procedures are provided in the Appendix~\ref{app:pipeline_figure}.

\subsection{Annotation Quality Assessment}

We evaluate annotation quality via human evaluation on 15 sampled videos across all annotation levels. We ask seven human experts to conduct evaluations on textual contents with visual contents being reference. Each video is rated by two annotators on a 1--5 Likert scale. As shown in Table~\ref{tab:human_eval_mms}, annotations achieve consistently high scores across levels, indicating strong accuracy, informativeness, and completeness. Inter-annotator agreement is moderate, reflecting the subjective nature of evaluating multimodal descriptions, particularly for saliency and linguistic quality. Additional details of the annotation quality human evaluation are included in Appendix~\ref{app:human_evaluation}.

\begin{figure}[t!]
  \centering
  \includegraphics[width=0.8\linewidth]{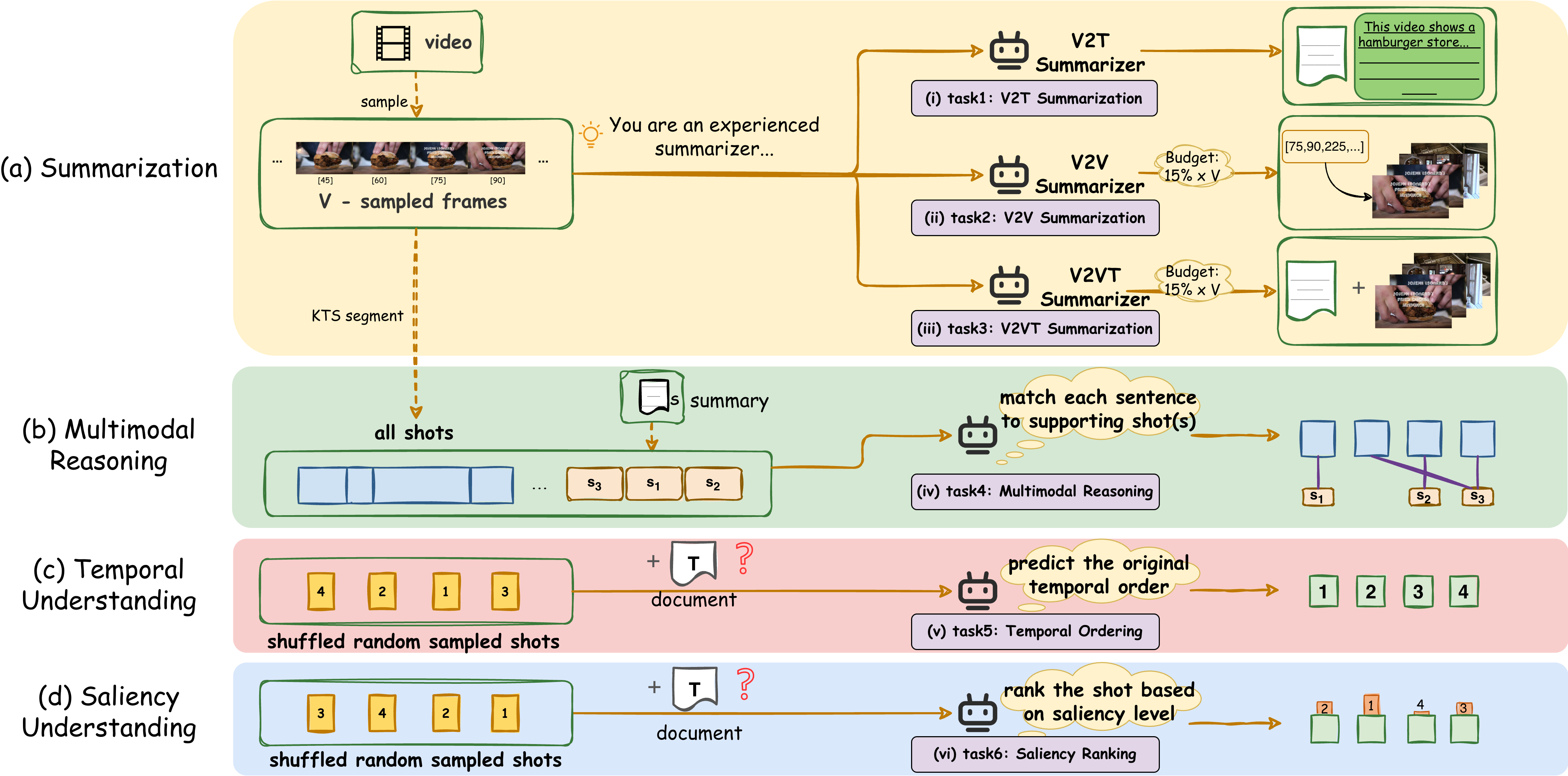}
  \caption{{Overview of tasks supported by HAVEN.}}
  \label{fig:experimental_setup}
  \vspace{-10pt}
\end{figure}
\subsection{Supporting Tasks}

Building on the aligned dataset, we define a unified suite of tasks to evaluate multimodal understanding in MLLMs, as illustrated in Figure~\ref{fig:experimental_setup}. These tasks cover complementary aspects of video understanding—including summarization, temporal reasoning, multimodal grounding, and saliency estimation. The tasks are also summarized below in Table~\ref{tab:task_definitions}.

\textbf{Multimodal Summarization.} HAVEN supports three multimodal summarization settings: \textbf{V2T (Video-to-Text)}, \textbf{V2V (Keyframe Selection)}, and \textbf{V2VT (Joint Video-Text) Summarization}. Given a video as input, the model is tasked with generating a textual summary, selecting a subset of keyframes, or performing both jointly. These tasks evaluate the model’s ability to identify important content from visual inputs and produce coherent summaries, detect visually salient content, and  maintain cross-modal consistency to ensure that generated summaries and selected keyframes correspond to the same underlying content.

\textbf{Temporal Understanding.} Given shuffled shots $\tilde{H}$, with optional document-level context $T$, the model predicts the original chronological order: ($\hat{\pi} = \mathcal{M}_{\text{temp}}(\tilde{H}, T)$), where $\hat{\pi}$ denotes a permutation over shots. This task evaluates the ability to reason about temporal dependencies and event progression.

\textbf{Multimodal Grounding.} Given shots $H$ and shuffled summary sentences $\tilde{S}$, the model predicts alignments $\hat{M} = \mathcal{M}_{\text{align}}(H, \tilde{S})$. $\hat{M} = \{(s_k, H_k)\}$ and $H_k \subseteq H$. This task evaluates fine-grained grounding by linking textual units to their corresponding visual evidence.

\textbf{Saliency Ranking.} Given sampled shots $\tilde{H}$, with or without document-level context $T$, the model predicts a saliency ranking ($\hat{\pi} = \mathcal{M}_{\text{rank}}(\tilde{H}, T)$). This task measures the ability to infer global importance and produce a coherent ranking of visual segments.

\section{Experiments}
\subsection{Evaluation Protocol}

We use the six tasks supported by HAVEN to comprehensively evaluate MLLMs. Each task is defined with its own input–output formulation, prompting protocol, and evaluation metrics. Due to the large scale of VideoXum and MR.HiSum, we evaluate on a randomly sampled subset of 100 videos from each dataset rather than the full benchmark. Detailed verbatim prompts for all tasks are provided in Appendix~\ref{app:prompts}. The overall evaluation results are summarized in Table~\ref{tab:overall_capability_weighted}.

\paragraph{Evaluation Metrics}

\textbf{(1) Summarization tasks}. We evaluate textual summaries using ROUGE~\citep{lin2004rouge}, BERTScore~\citep{zhang2019bertscore}, and a simplified G-Eval-style LLM-as-a-judge metric~\citep{GEval2023Liu}. Keyframe selection is evaluated using F1 against human-annotated keyframes. Cross-modal alignment is evaluated using CLIPScore~\citep{hessel2021clipscore}. All models take the full set of sampled frames as input and are required to select keyframes under a fixed 15\% budget following ~\citet{zhang2016video}. \textbf{(2) Temporal and saliency understanding tasks}. We evaluate performance using rank correlation metrics, including Spearman’s $\rho$~\citep{spearman1987proof} and Kendall’s $\tau$~\citep{kendall1938new}. \textbf{(3) Multimodal grounding task}. We evaluate using F1 score as the primary metric to evaluate the overlap between predicted and ground-truth shots, with IoU~\citep{rezatofighi2019generalized} and Hit@1 reported for additional analysis.

\paragraph{Evaluation Models}
We evaluate both commercial and open-source models, including GPT-5.2~\citep{GPT5}, Qwen3-VL-8B-Instruct~\citep{Qwen3VL2025bai}, Qwen2.5-VL-7B-Instruct~\citep{Qwen2.5VL}, InternVideo2.5-Chat-8B~\citep{InternVideo252025wang}, VideoLLaMA3-7B~\citep{zhang2025videollama}, and GLM-4.6V-Flash~\citep{GLM4.6V}, in a zero-shot setting. Task-specific settings are described in the corresponding subsections.

\begin{table*}[t]
\caption{Unified results across HAVEN task families. Bold indicates the best performance in each column. “--” indicates that the model is not evaluated under the corresponding setting, either because the w/ Doc setting is not applied (for text-proxy LLMs) or due to input frame limitations (e.g., GPT-5.2 and GLM in summarization).}
\centering
\small
\setlength{\tabcolsep}{3pt}
\renewcommand{\arraystretch}{1.0}
\resizebox{\textwidth}{!}{
\begin{tabular}{lcccc|cc|c|cc}
\hline
\multirow{2}{*}{\textbf{Model}}
& \multicolumn{4}{c}{\textbf{Summarization}}
& \multicolumn{2}{c}{\textbf{Temporal}}
& \multicolumn{1}{c}{\textbf{Grounding}}
& \multicolumn{2}{c}{\textbf{Saliency}} \\
\cline{2-10}
& \textbf{V2T. BS}
& \textbf{V2V. F1}
& \textbf{V2VT. BS}
& \textbf{V2VT. F1}
& \textbf{w/o Doc $\rho$}
& \textbf{w/ Doc $\rho$}
& \textbf{F1}
& \textbf{w/o Doc $\rho$}
& \textbf{w/ Doc $\rho$} \\
\hline
text-proxy LLM& 0.856 & 0.171 & 0.858 & 0.171 & 0.019 & --& \textbf{0.804} & 0.154 & --\\
GPT-5.2~(\citeyear{GPT5})& -- & -- & -- & -- & \textbf{0.280} & \textbf{0.461} & 0.510 & \textbf{0.210} & 0.190 \\
GLM-4.6V~(\citeyear{GLM4.6V})    & -- & -- & -- & -- & 0.147 & 0.306 & 0.457 & 0.164 & 0.136 \\
InternVideo2.5~(\citeyear{InternVideo252025wang})    & 0.847 & 0.273 & 0.859 & 0.065 & 0.083 & 0.089 & 0.394 & 0.169 & 0.178\\ 
Qwen3-VL~(\citeyear{Qwen3VL2025bai})       & 0.865 & 0.173 & 0.859 & 0.235 & 0.230 & 0.300 & 0.443 & 0.187 & \textbf{0.218} \\
Qwen2.5-VL~(\citeyear{Qwen2.5VL})     & \textbf{0.874}& 0.209 & 0.866 & 0.176 & 0.068 & 0.088 & 0.299 & 0.006 & 0.070 \\
VideoLLaMA3~(\citeyear{zhang2025videollama})    & \textbf{0.874} & \textbf{0.293} & \textbf{0.869} & \textbf{0.267} & 0.053 & 0.064 & 0.253 & 0.142 & 0.164 \\
\hline
\end{tabular}
}
\label{tab:overall_capability_weighted}
\end{table*}

\subsection{Summarization Tasks Results}

We evaluate summarization from three settings. Results are reported in Table~\ref{tab:exp123_detailed_weighted}. As a text-proxy LLM baseline, GPT-5.4~\citep{GPT5} operates on frame descriptions and follows the same budget. We note that GPT-5.2 and GLM models impose strict limits on the number of input frames, preventing them from joining summarization tasks.

In the V2T setting, VideoLLaMA3 and Qwen2.5-VL achieve the highest or near-highest scores across metrics, while InternVideo2.5 performs worse than the baseline with     aggregated ROUGE score of 0.152 and G-Eval score below 2.5. In the V2V setting, all models outperform the baseline. VideoLLaMA3 achieves the best performance (0.292 on precision and 0.319 on recall), closely followed by InternVideo2.5. Both MLLMs and the text-proxy LLM achieve keyframe selection F1 scores in the range of \textit{0.15--0.30}, which are substantially above random selection, yet remain far from satisfactory, highlighting the limited capability of current models in accurately selecting key visual information. In the joint V2VT setting, we compare joint-generation CLIPScore with the alignment scores from independently generated V2T and V2V outputs. The alignment is not consistently improved. 

The text-proxy LLM shows stable performance across all settings, with similar positions of its datapoints across the two subfigures in Fig.~\ref{fig:results_exp123_summarization}. In contrast, most MLLMs exhibit degraded performance under the joint generation setting, reflected by the downward shift of their datapoints in the right subfigure. VideoLLaMA3 is a notable exception, remaining relatively stable and  achieving strong performance in all tasks. InternVideo2.5 improves in text summarization under the joint setting, but suffers a substantial decline in frame selection performance, suggesting a trade-off between textual generation and visual content selection. Qwen models generally exhibit decreased text summarization performance under joint generation, while Qwen3-VL shows a notable improvement in keyframe selection. In both settings, BERTScore differences across models are marginal (between \textit{0.84--0.88}), indicating that textual summarization quality is largely comparable across models and is not substantially affected by the generation settings. 

{Overall}, the three tasks reveal a consistent pattern in model performance. Current MLLMs are generally strong at producing fluent textual summaries but show weaker performance in keyframe selection. Joint multimodal generation does not consistently improve alignment between textual and visual outputs. 

\begin{table*}[t]
\caption{Detailed results for summarization experiments under a unified evaluation setting. 
$\Delta$ denotes the change in CLIPScore relative to the corresponding V2T–V2V alignment. 
G-Eval refers to a GPT-based evaluation score following the G-Eval(\citeyear{GEval2023Liu}) protocol. 
ROUGE denotes the aggregated score of ROUGE-1, ROUGE-2, and ROUGE-L.}
\centering
\small
\setlength{\tabcolsep}{3pt}
\renewcommand{\arraystretch}{1.02}
\resizebox{0.72\textwidth}{!}{
\begin{tabular}{lcc|cc|ccccc}
\hline
\multirow{2}{*}{\textbf{Model}} & \multicolumn{2}{c}{\textbf{V2T. Sum}} & \multicolumn{2}{c}{\textbf{V2V. Sum}} & \multicolumn{5}{c}{\textbf{V2VT. Sum}} \\
\cline{2-3}\cline{4-5}\cline{6-10}
 & \textbf{G-Eval} & \textbf{ROUGE} & \textbf{P} & \textbf{R} & \textbf{G-Eval} & \textbf{ROUGE} & \textbf{P} & \textbf{R} & \textbf{CLIPScore ($\Delta$)} \\
\hline
text-proxy LLM & \textbf{3.528} & 0.203 & 0.172 & 0.178 & 3.481 & \textbf{0.213} & 0.173 & 0.178 & \textbf{0.899} {\scriptsize(-0.024)} \\
InternVideo2.5 & 2.252 & 0.152 & 0.269 & 0.299 & 3.149 & 0.160 & \textbf{0.433} & 0.038 & 0.741 {\scriptsize(-0.077)} \\
Qwen3-VL & 3.412 & 0.207 & 0.186 & 0.182 & 3.380 & 0.195 & 0.204 & 0.481 & 0.881 {\scriptsize(-0.020)} \\
Qwen2.5-VL & 3.372 & \textbf{0.235} & 0.211 & 0.229 & 3.478 & 0.197 & 0.200 & 0.249 & 0.846 {\scriptsize(-0.013)} \\
VideoLLaMA3 & 3.353 & 0.223 & \textbf{0.292} & \textbf{0.319} & \textbf{3.536} & 0.212 & 0.218 & \textbf{0.558} & 0.850 {\scriptsize(+0.044)} \\
\hline
\end{tabular}
}
\label{tab:exp123_detailed_weighted}
\end{table*}

\subsection{Temporal Understanding Task Results}

We evaluate temporal understanding by asking models to recover the chronological order of shuffled video shots. Each shot is represented by a subset of sampled frames. We consider two input settings: a \textit{w/o Doc} setting that provides only the shuffled shots, and a \textit{w/ Doc} setting that additionally provides the full reference document derived from HAVEN dataset. As a text-proxy baseline, we replace each shot with its description and ask GPT-5.4 to recover the temporal order from text.

\begin{figure*}[b]
    \centering

    \begin{subfigure}[t]{0.3\textwidth}
        \centering
        \includegraphics[width=\linewidth]{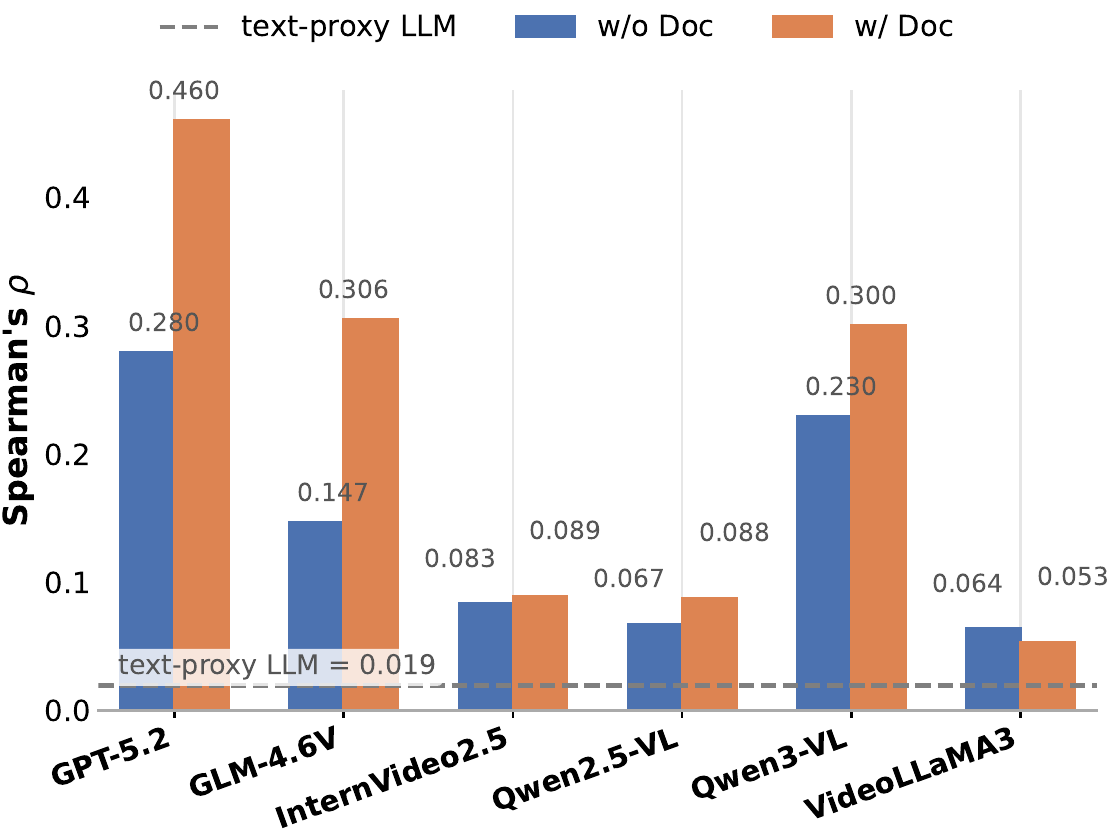}
        \caption{Temporal understanding model comparison.}
        \label{fig:exp4_a}
    \end{subfigure}
    \hfill
    \begin{subfigure}[t]{0.38\textwidth}
        \centering
        \includegraphics[width=\linewidth]{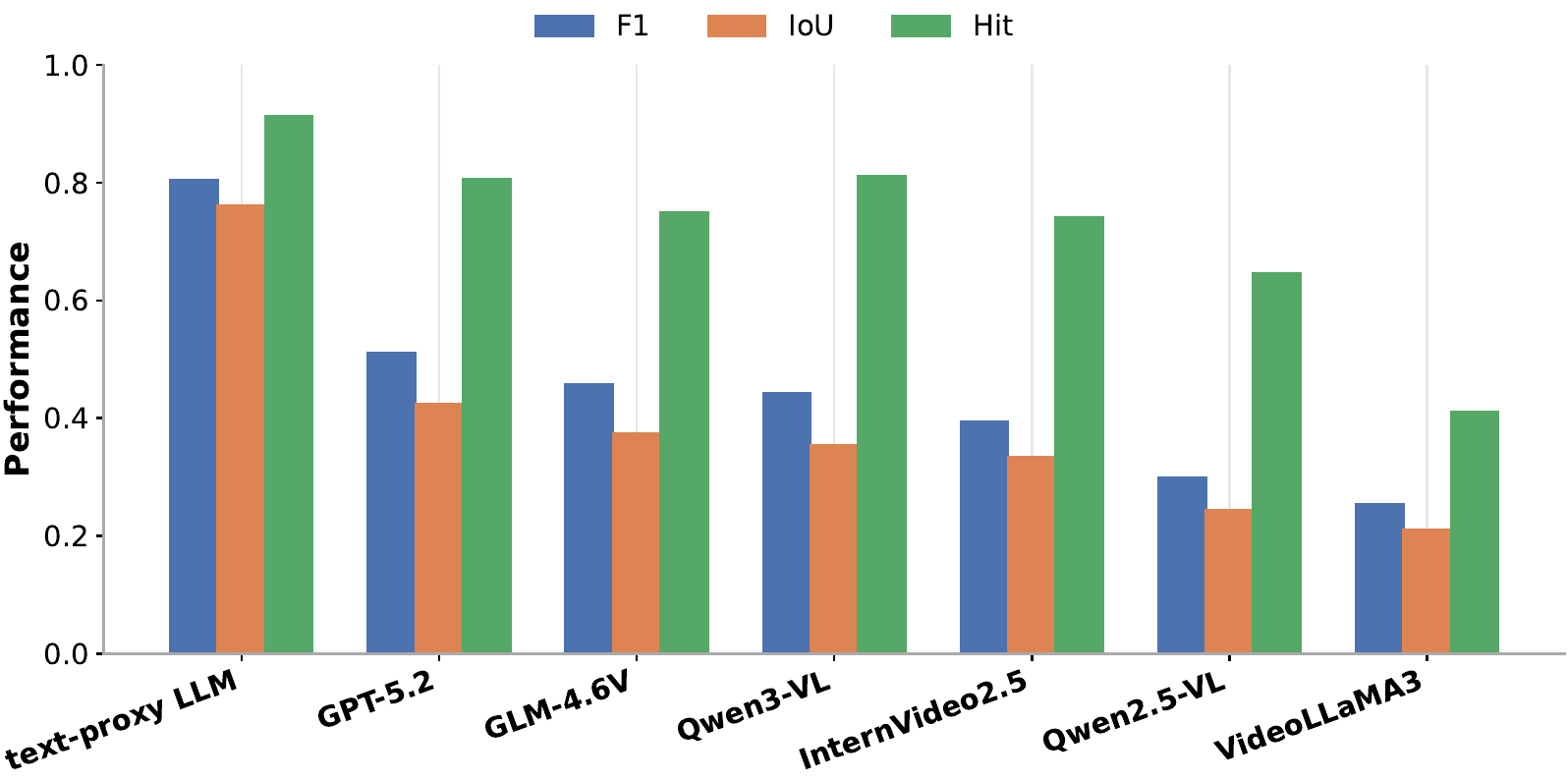}
        \caption{Multimodal grounding model comparison. The text-proxy LLM consistently outperforms multimodal models across metrics.}
        \label{fig:results_exp5}
    \end{subfigure}
    \hfill
    \begin{subfigure}[t]{0.3\textwidth}
        \centering
        \includegraphics[width=\linewidth]{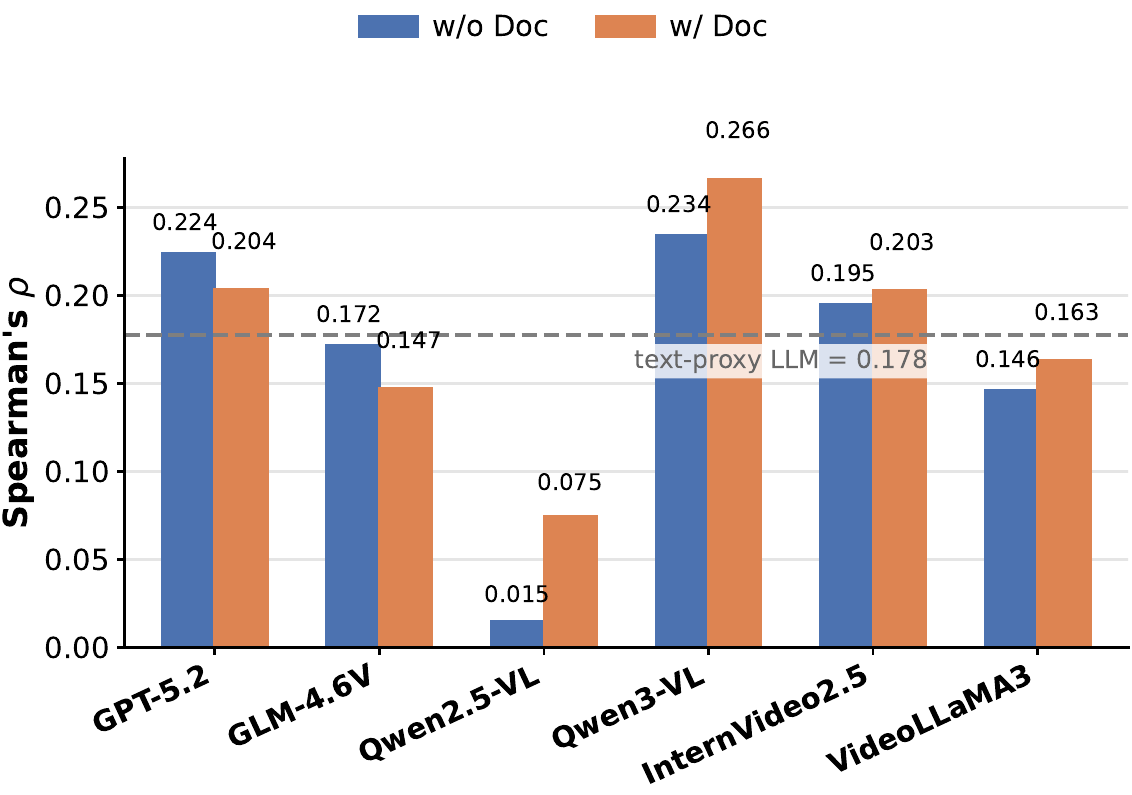}
        \caption{Saliency ranking performance. }
        \label{fig:exp6_bars}
    \end{subfigure}
    \caption{Comparison across temporal understanding, multimodal grounding, and saliency ranking tasks. The dashed horizontal line indicates the text-proxy baseline.}
    \label{fig:exp456_combined}
\end{figure*}

Fig.~\ref{fig:exp4_a} shows that the text-proxy LLM performs substantially worse than all multimodal models, suggesting text-only is insufficient for temporal reasoning. Additionally, providing the full document generally improves performance across models, with GPT-5.2 and GLM-4.6V benefiting the most(\textit{+108\%} and \textit{+64.7\%}, respectively) and Qwen3-VL showing stable gains. An exception is observed for VideoLLaMA3, where the \textit{w/ Doc} setting leads to a negligible performance drop, suggesting potential difficulty in effectively integrating additional context. Across all models, GPT-5.2 achieves the best overall performance, while Qwen3-VL performs best among open-source models. 

\subsection{Multimodal Grounding Task Results}

In inputs, each shot is represented by uniformly sampled frames. As a text-proxy baseline, we provide GPT-5.4 with shot-level descriptions and shuffled summary sentences to perform grounding based on text alone.

Overall, the text-only baseline achieves strong performance, outperforming most multimodal models. Across metrics, we observe a consistent pattern: Hit@1 > F1 > IoU, as correctly identifying at least one relevant segment is easier than achieving high overlap with the ground-truth alignment. Commercial model GPT-5.2 achieves best scores across metrics.

Among open-source models in Fig.~\ref{fig:results_exp5}, Qwen3-VL and GLM-4.6V show relatively strong performance, while VideoLLaMA3 performs lag far behind by all models. Notably, GLM-4.6V achieves higher F1 and IoU than Qwen3-VL, but lower Hit@1 (green bar). This indicates that GLM tends to produce more precise but less robust alignments, whereas Qwen3-VL more reliably identifies relevant segments but with lower overlap quality.

\subsection{Saliency Ranking Task Results}

In the detailed experiment, we consider two settings: a \textit{w/o Doc} setting using visual input only, and a \textit{w/ Doc} setting that additionally provides the full video document. As a text-only baseline, GPT-5.4 ranks shots based on shot-level descriptions.

Model performance varies substantially across saliency ranking. Comparing the \textit{w/o Doc} and \textit{w/ Doc} settings shown with blue bars and orange bars on Fig.~\ref{fig:exp6_bars}, providing document-level context generally improves performance. Qwen2.5-VL consistently underperforms across models with a large gap on bars, while Qwen3-VL achieves the best overall performance, outperforming both open-source and commercial models. This discrepancy may be influenced by the task formulation rather than reflecting general understanding capability. The text-proxy LLM achieves mid-level performance, indicating that textual information alone provides useful but insufficient signals for saliency estimation.

Notably, saliency awareness is a fundamental component of summarization, yet we observe a clear gap between performance on summarization and saliency ranking tasks in Fig.~\ref{fig:overall_radar}. For example, VideoLLaMA3 performs strongly on summarization but achieves only moderate performance on saliency ranking. In contrast, Qwen3-VL exhibits relatively stable performance across tasks and covers the largest area in the radar plot, indicating more balanced multimodal understanding. These results highlight that strong summarization performance does not necessarily imply accurate saliency understanding.

\section{Analysis}
\subsection{Cross-task Analysis}

\begin{wrapfigure}{r}{0.5\linewidth}
  \centering
 \includegraphics[width=\linewidth]{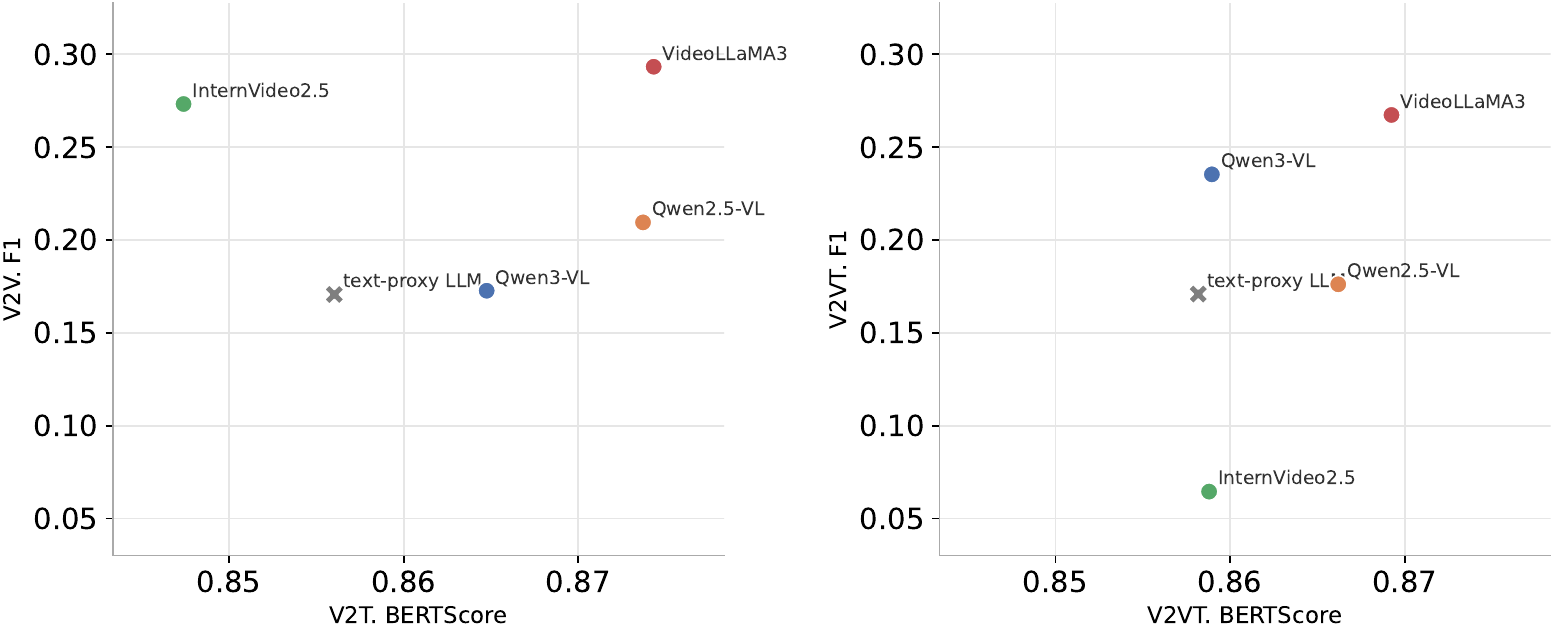}
  \caption{Summarization performance relationships across tasks. (a) text summarization quality (BERTScore) vs. keyframe selection performance (F1). 
(b) joint summarization quality.}
\label{fig:results_exp123_summarization}
\end{wrapfigure}

We observe that model performance is highly sensitive to task formulation, suggesting that multimodal capabilities cannot be reliably assessed using a single evaluation setup. For example, while some models achieve strong performance on textual summarization (V2T)in Fig.~\ref{fig:results_exp123_summarization}, their performance drops substantially on tasks that require explicit visual grounding, such as keyframe selection (V2V) and multimodal grounding (Fig.~\ref{fig:overall_radar}). This indicates that high-quality summaries may rely more on language generation ability than on accurate visual understanding. 

A similar discrepancy is observed between summarization and saliency ranking. As illustrated in the overall radar plot Fig.~\ref{fig:overall_radar}, model performance on summarization is highly concentrated, with most models achieving similar results. In contrast, performance on saliecny ranking exhibits substantially clearer separation across models, despite the conceptual similarity between the two tasks. This suggests that summarization reflects a more holistic and implicit notion of importance, whereas saliency ranking requires more explicit modeling of visual importance, making it a more direct test of visual understanding.


These findings highlight the need for multi-faceted evaluation, as different task formulations probe different aspects of multimodal capability and may lead to substantially different conclusions about model performance.

\subsection{Multimodal Fusion Analysis}

\begin{wrapfigure}{r}{0.4\linewidth}
      \includegraphics[width=\linewidth]{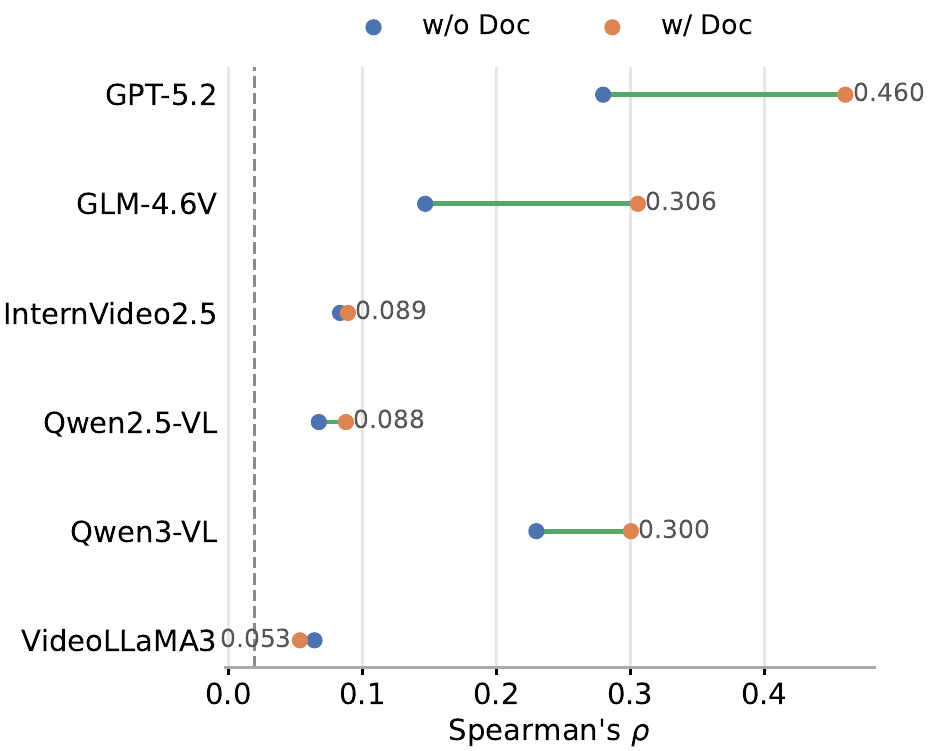}
      \caption{Temporal understanding model gains from multimodal input.}
      \label{fig:exp4_gap}
\end{wrapfigure}

Across multiple tasks, we observe a consistent performance trend in Figs.~\ref{fig:exp4_a} and ~\ref{fig:exp6_bars}, that is multimodal inputs generally outperform visual-only inputs, which in turn outperform text-only inputs (which is represented by the texy-procy baseline). This indicates that models are able to benefit from the complementary information provided by multiple modalities. However, this advantage does not extend to cross-modal alignment. In multimodal grounding (Fig.~\ref{fig:results_exp5}), models perform significantly better when aligning within the same modality than across modalities, where the texy-proxy LLM bars reaching far more higher on all metrics. 

These results suggest that while current MLLMs can effectively utilize multimodal inputs for task-specific predictions, they struggle to establish precise semantic correspondences between modalities. In other words, models can leverage multimodal information for decision making, but lack the ability to reliably align and integrate representations across modalities.



\section{Conclusions}
In this work, we present \textbf{HAVEN}, a novel hierarchically aligned multimodal benchmark for comprehensively evaluating the capabilities of multimodal models on video understanding across multimodal summarization, temporal understanding, grounding, and saliency understanding. Our zero-shot experiments reveal a gap between surface-level textual fluency and grounded multimodal understanding. We hope HAVEN can serve as a standardized benchmark to facilitate the development of advanced multimodal models that not only generate coherent text, but also effectively ground, align and reason over both textual and visual information in a unified manner.




\section*{Limitations}
While HAVEN provides a unified framework for evaluating multimodal video understanding, it has several limitations that open directions for future work. (1) Our benchmark is constructed by extending existing video summarization datasets with hierarchical annotations and cross-modal alignments, rather than collecting a fully new dataset from scratch. Although this design enables scalable construction and controlled evaluation, it may inherit biases and limitations from the original datasets. (2) HAVEN focuses on two modalities (video and text), while omitting other potentially informative modalities such as audio. Prior work suggests that audio signals can provide complementary cues for semantic understanding. Incorporating additional modalities is an important direction for improving the completeness of multimodal evaluation. (3) Our current benchmark primarily targets relatively short videos. Extending HAVEN to longer-form videos would enable the study of long-range video understanding and memory, which remain challenging for current MLLMs. We leave these directions for future work.

\bibliographystyle{plainnat}
\bibliography{main}






\newpage
\appendix
\renewcommand{\thetable}{A\arabic{table}}
\renewcommand{\thefigure}{A\arabic{figure}}
\setcounter{table}{0}
\setcounter{figure}{0}
\section{Dataset Construction Pipeline}
\label{app:pipeline_figure}

\subsection{Details of Pipeline}
This section describes how HAVEN Benchmark dataset is constructed from raw source videos and some benchmark metadata into a unified hierarchical annotation format. Starting from sampled video frames and benchmark-provided structural signals, we progressively rebuilt the structural signals and build frame-level descriptions, shot-level descriptions, video-level documents, concise summaries, and sentence-to-shot alignments. The resulting data format supports evaluation at multiple granularities while preserving explicit alignments between visual content and textual annotations.

\begin{figure*}[t]
\centering
\includegraphics[width=\textwidth]{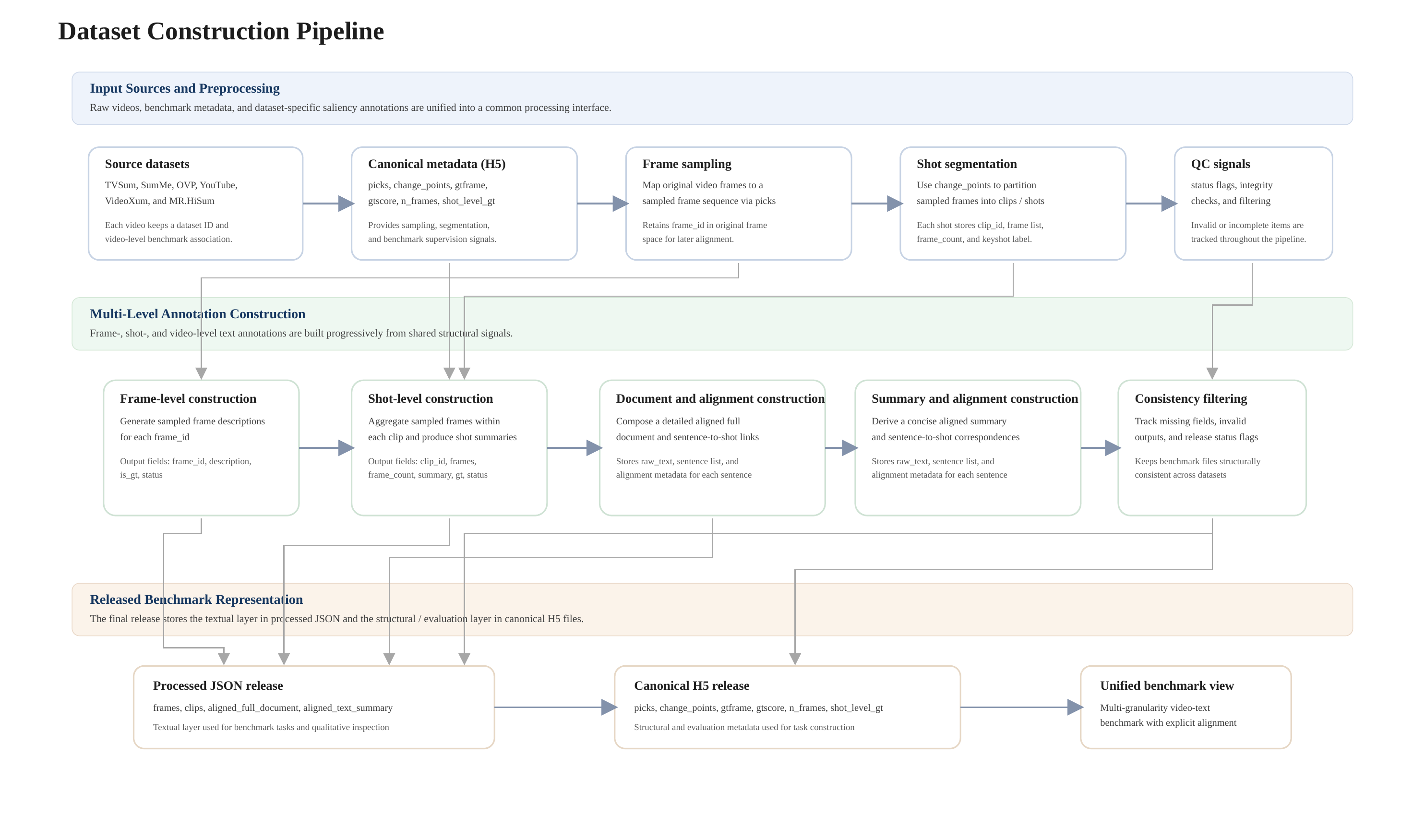}
\caption{Overview of the HAVEN Benchmark dataset construction pipeline.}
\label{fig:dataset_construction_pipeline}
\end{figure*}

\subsection{HAVEN Benchmark Data Sample}
\label{app:dataset_sample}

\begin{figure*}[htbp]
\centering
\includegraphics[width=\textwidth]{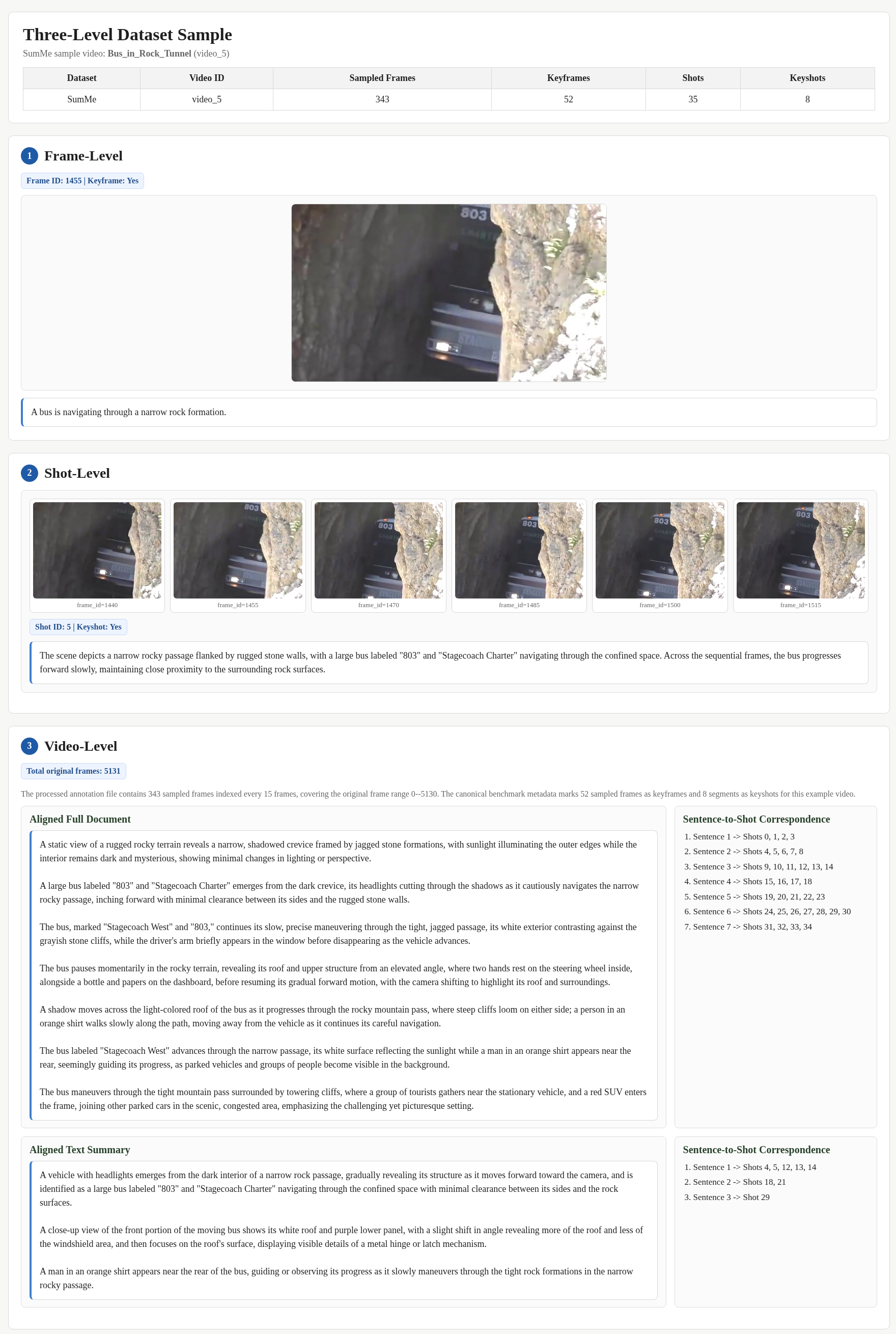}
\caption{A three-level hierarchically annotated example from HAVEN dataset, shown with a video from SumMe (\texttt{video\_5}, \textit{Bus\_in\_Rock\_Tunnel}). }
\label{fig:dataset_sample_three_level}
\end{figure*}

We show a sample from our hierarchically aligned dataset HAVEN in Fig.~\ref{fig:dataset_sample_three_level}. The example shows vidual and textual information included in a video data and demonstrates how visual contents align with textual units across levels. This hierarchical organization enables unified evaluation across multiple tasks in HAVEN benchmark.

\section{Human Evaluation of Annotation Quality}
\label{app:human_evaluation}
To further validate the quality of the annotations in MMS-Benchmark, we conduct a human evaluation on a subset of benchmark videos. The evaluation is designed to assess annotation quality at multiple granularities, including frame-level descriptions, shot-level descriptions, full-document annotations, and summary annotations.
\subsection{Annotation Setup}
We perform human evaluation on a subset of 15 videos sampled from MMS-Benchmark. Each video is independently evaluated by two human annotators using a 1--5 Likert scale. For each video, annotators are asked to assess: (1) two sampled frame-level annotations, (2) one sampled shot-level annotation, (3) the corresponding full-document annotation, and (4) the corresponding summary annotation.

Following our evaluation protocol, frame- and shot-level assessment is conducted on a fixed subset predefined in the evaluation set. For consistency, when a legacy submission contains denser frame- or shot-level judgments, we only retain the protocol-aligned subset for aggregation. Annotators are provided with the video content together with the corresponding annotations and are instructed to score each item independently according to predefined rubrics. The final score for each metric is computed by averaging the scores from the two annotators on each video, and then reporting the mean and standard deviation across all evaluated videos.

\subsection{Evaluation Dimensions}
We evaluate annotation quality at four levels.
\paragraph{Frame-level annotation.}
Each sampled frame description is evaluated on: (1) \textbf{Accuracy}: whether the description faithfully reflects the visible content in the frame; (2) \textbf{Informativeness}: whether the description captures sufficient and useful visual details.

\paragraph{Shot-level annotation.}
Each sampled shot description is evaluated on:
(1) \textbf{Accuracy}: whether the description correctly summarizes the visual event or content of the shot; (2) \textbf{Informativeness}: whether the description conveys the key information contained in the shot.

\paragraph{Full-document annotation.}
Each full-document annotation is evaluated on: (1) \textbf{Accuracy}: whether the document is factually consistent with the video; (2) \textbf{Completeness}: whether the document adequately covers the major events and salient content in the video; (3) \textbf{Readability}: whether the document is coherent, fluent, and easy to understand.

\paragraph{Summary annotation.}
Each summary annotation is evaluated on: (1) \textbf{Accuracy}: whether the summary is faithful to the video content; (2) \textbf{Relevance}: whether the summary focuses on the most important content; (3) \textbf{Conciseness}: whether the summary avoids redundancy while remaining informative; (4) \textbf{Readability}: whether the summary is fluent and easy to read.

\subsection{Annotation Protocol \& Rubrics}
All metrics are rated on a 1--5 Likert scale, where a higher score indicates better quality.

\paragraph{General rubric.}
\begin{itemize}
    \item \textbf{5}: excellent; fully satisfies the criterion with no noticeable issues;
    \item \textbf{4}: good; mostly satisfies the criterion with only minor issues;
    \item \textbf{3}: acceptable; partially satisfies the criterion but has clear room for improvement;
    \item \textbf{2}: poor; fails to satisfy the criterion in several important aspects;
    \item \textbf{1}: very poor; largely fails to satisfy the criterion.
\end{itemize}

Annotators are instructed to base their judgments only on the correspondence between the video and the annotation, rather than on stylistic preference. They are also asked to score each metric independently. For example, an annotation may be highly accurate but not sufficiently informative, or readable but incomplete. To quantify inter-annotator consistency, we report quadratic weighted Cohen's $\kappa_w$, Spearman correlation $\rho$, and exact agreement rate.

\begin{figure*}[t]
\centering
\includegraphics[width=0.6\textwidth]{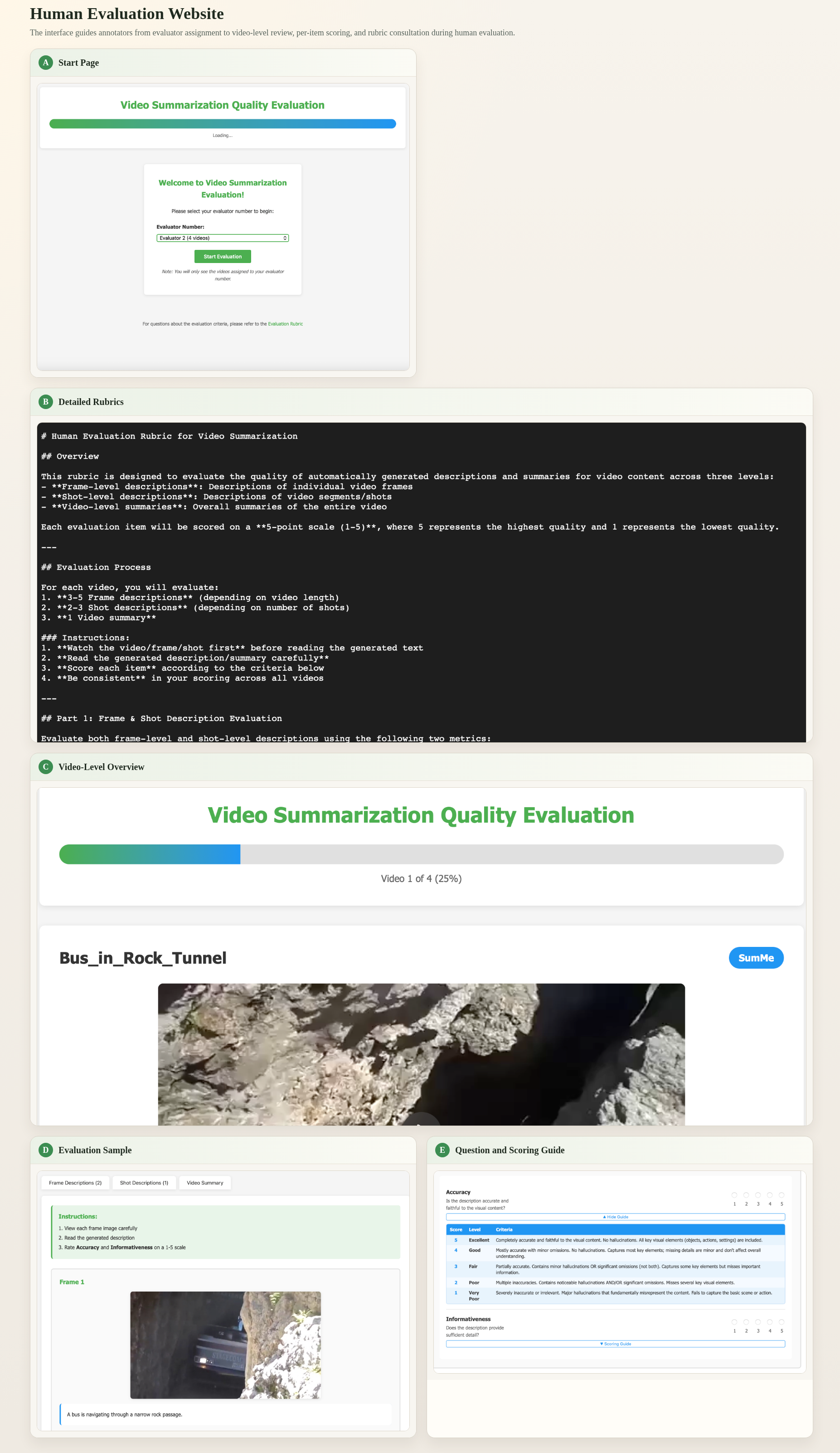}

\caption{Screenshots of the human evaluation website used to assess HAVEN dataset annotations. Panel A shows the start page. Panel B shows the detailed rubric page. Panel C--E shows the main evaluation interface. }
\label{fig:human_eval_interface}
\end{figure*}

\paragraph{Website interface and rubric presentation.}
Figure~\ref{fig:human_eval_interface} shows the website used in our human evaluation. The interface is organized as a structured workflow. Within the evaluation view, the website presents both global and local evidence for judgment. Each block asks annotators to rate a single quality dimension on a 1--5 Likert scale, thereby separating different judgment criteria. The rubric is integrated into the website in two complementary forms: (1) A standalone rubric page provides the full definitions and score interpretations for all evaluation dimensions. (2) The scoring interface itself exposes expandable scoring guides next to each question, allowing annotators to consult the relevant criteria at the point of judgment.

\section{Prompt Design and Templates}
\label{app:prompts}

\begin{scriptsize}
\setlength{\LTleft}{0pt}
\setlength{\LTright}{0pt}
\setlength{\tabcolsep}{4pt}
\renewcommand{\arraystretch}{1.08}

\begin{longtable}{p{1.65cm}p{2cm}p{2cm}p{3.00cm}p{3.00cm}}

\caption{Verbatim prompt templates used in HAVEN experiments. Main task prompts and auxiliary text-proxy LLM settings are listed separately.}
\label{tab:prompt_summary_appendix}\\

\hline
\textbf{Task} & \textbf{Setting} & \textbf{Model Input} & \textbf{Prompt Template (Verbatim)} & \textbf{Required Output} \\
\hline
\endfirsthead

\hline
\textbf{Task} & \textbf{Setting} & \textbf{Model Input} & \textbf{Prompt Template (Verbatim)} & \textbf{Required Output} \\
\hline
\endhead

\multicolumn{5}{c}{\textit{Main experiments used in the paper}} \\
\hline
Summarization & V2T & Chronological video frames
& ``You are given $N$ frames from a video shown in chronological order.'' ``Please write a concise summary of the video, describing the main events, actions, scenes, and topics from beginning to end.'' ``Your summary should be 3--5 sentences.''
& Free-form summary in 3--5 sentences. \\
\hline

Summarization & V2V & Chronological video frames
& ``Below are $N$ frames from a video in chronological order:'' ``From these $N$ frames, select the $M$ most important frames that best represent the key moments of the video.'' ``Output ONLY a JSON array of frame numbers (1-indexed integers), with exactly $M$ elements. Example: [1, 5, 12]'' ``Do not include any explanation.''
& JSON array of 1-indexed frame IDs, e.g., ``[1, 5, 12]''. \\
\hline

Summarization & V2VT & Chronological video frames
& ``Based on all $T$ frames above, complete TWO tasks:'' ``TASK 1 --- Write a summary of this video in EXACTLY $K$ sentences.'' ``TASK 2 --- Select EXACTLY $M$ most important frames (no more, no less) from: Frame 1, Frame 2, ..., Frame $T$.'' ``Use EXACTLY this format (no extra text): SUMMARY: 1. [sentence 1] ... IMPORTANT FRAMES: [Frame X, Frame Y, ...]''

& ``SUMMARY:'' followed by numbered sentences, then ``IMPORTANT FRAMES: [Frame X, Frame Y, ...]''. \\
\hline

Temporal understanding & w/o Doc & Shuffled shots with frames
& ``Below are $N$ shots from a video, presented in shuffled order:'' ``You have been shown $N$ shots (Shot 1, Shot 2, ..., Shot $N$), presented in shuffled order.'' ``Reorder ALL shots to match the actual temporal sequence of the video, from first to last.'' ``Output ONLY the ordering in this exact format (use > between shots): Shot X > Shot Y > Shot Z > ... Include all shots exactly once.''

& ``Shot X > Shot Y > ...'' with all shots exactly once. \\
\hline

Temporal understanding & w/ Doc & Shuffled shots with frames; full-document context
& ``Below are $N$ shots from a video, presented in shuffled order:'' ``The video's full text description: [DOC]'' ``You have been shown $N$ shots (Shot 1, Shot 2, ..., Shot $N$), presented in shuffled order.'' ``Reorder ALL shots to match the actual temporal sequence of the video, from first to last.'' ``Output ONLY the ordering in this exact format (use > between shots): Shot X > Shot Y > Shot Z > ... Include all shots exactly once.''

& ``Shot X > Shot Y > ...'' with all shots exactly once. \\
\hline

Multimodal grounding & w/o Doc & Video clips and shuffled summary sentences
& ``Below are $C$ clips from a video, labeled Clip 1 through Clip $C$:'' ``Here are $S$ sentences from a text summary of this video (the order has been shuffled):'' ``For each summary sentence, identify which clip(s) it corresponds to.'' ``Use exactly this format (list clip numbers separated by commas): Summary 1: Clip X, Clip Y''

& One line per sentence: ``Summary $i$: Clip X, Clip Y''. \\
\hline

Saliency ranking & w/o Doc & Selected shots with frames
& ``Below are $N$ shots from the video, labeled Shot 1 through Shot $N$:'' ``You have seen $N$ shots from a video (Shot 1, Shot 2, ..., Shot $N$).'' ``Rank ALL of these shots from most important to least important based on their visual content and relevance to the overall video.'' ``Output ONLY the ranking in this exact format (use > between shots): Shot X > Shot Y > Shot Z > ... Include all shots exactly once.''
& ``Shot X > Shot Y > ...'' with all shots exactly once. \\
\hline

Saliency ranking & w/ Doc & Selected shots with frames and document context
& ``Background document describing the video: [DOC]'' ``Below are $N$ shots from the video, labeled Shot 1 through Shot $N$:'' ``You have seen $N$ shots from a video (Shot 1, Shot 2, ..., Shot $N$).'' ``Rank ALL of these shots from most important to least important based on their visual content and relevance to the overall video.'' ``Output ONLY the ranking in this exact format (use > between shots): Shot X > Shot Y > Shot Z > ... Include all shots exactly once.''
& ``Shot X > Shot Y > ...'' with all shots exactly once. \\
\hline

\multicolumn{5}{c}{\textit{Text-proxy LLM baselines used for comparison}} \\
\hline
Summarization & V2VT text-proxy & Chronological frame descriptions only
& ``Below are text descriptions of $N$ sampled frames from a video, listed in chronological order as Frame 1 to Frame $N$:'' ``Based on the frame descriptions above, complete BOTH tasks below.'' ``TASK 1 --- VIDEO SUMMARY:'' ``TASK 2 --- IMPORTANT FRAMES:'' ``IMPORTANT FRAMES: [X, Y, ...]''

& ``SUMMARY:'' plus numbered sentences; ``IMPORTANT FRAMES: [X, Y, ...]''. \\
\hline

Summarization & V2T text-proxy & Chronological frame descriptions only
& ``Below is a chronological list of $N$ frame-level descriptions sampled from a video.'' ``Each line describes one sampled frame. Use the full sequence to infer the overall video content.'' ``Based on the frame descriptions above, complete the task below.'' ``TASK --- VIDEO SUMMARY:'' ``SUMMARY:''
& ``SUMMARY:'' followed by numbered sentences only. \\
\hline

Summarization & V2V text-proxy & Chronological frame descriptions only
& ``Below is a chronological list of $N$ frame-level descriptions sampled from a video.'' ``Each line describes one sampled frame.'' ``Select exactly $M$ frame number(s) that best capture the key content and events of the video.'' ``Output ONLY a JSON array of 1-indexed integers with exactly that many elements.'' ``Example: [1, 5, 12]'' ``Do not include any explanation.''

& JSON array of 1-indexed frame IDs. \\
\hline

Temporal understanding & w/o Doc text-proxy & Shuffled event sentences only
& T``Below are $N$ sentences describing events in a video, shown in shuffled order:'' ``You have been shown $N$ sentences (Sentence 1, Sentence 2, ..., Sentence $N$), presented in shuffled order.'' ``Reorder ALL sentences to match the actual temporal sequence of the video, from first to last.'' ``Output ONLY the ordering in this exact format (use > between sentences): Sentence X > Sentence Y > Sentence Z > ... Include all sentences exactly once.''
& ``Sentence X > Sentence Y > ...'' with all sentences exactly once. \\
\hline

Grounding & w/o Doc text-proxy & Clip descriptions and shuffled summary sentences
& ``Below are text descriptions of $N$ clips from a video, labeled Clip 1 through Clip $N$:'' ``Here are $S$ sentences from a text summary of this video (the order has been shuffled):'' ``For each summary sentence, identify which clip(s) it corresponds to.'' ``Use exactly this format (list clip numbers separated by commas): Summary 1: Clip X, Clip Y''
& One line per sentence: ``Summary $i$: Clip X, Clip Y''. \\
\hline

Saliency ranking & w/o Doc text-proxy & Shot descriptions only
& ``Below are $N$ shots from the video, labeled Shot 1 through Shot $N$:'' ``You have read descriptions of $N$ shots from a video (Shot 1, Shot 2, ..., Shot $N$).'' ``Rank ALL of these shots from most important to least important based on their content and relevance to the overall video.'' ``Output ONLY the ranking in this exact format (use > between shots): Shot X > Shot Y > Shot Z > ... Include all shots exactly once.''
& ``Shot X > Shot Y > ...'' with all shots exactly once. \\
\hline

\end{longtable}
\end{scriptsize}




\end{document}